%% file: main.tex
\documentclass[sigconf,authorversion,nonacm]{acmart}
\usepackage{tikz}
\usepackage{xspace}
\usepackage{caption}
\usetikzlibrary{positioning,arrows.meta,calc,fit,shadows,automata,shapes, calc, arrows, backgrounds}

\AtBeginDocument{%
  }

\newcommand{\eg}{\textit{e.g.}\xspace}

\newcommand{\ie}{\textit{i.e.}\xspace}

\newcommand{\demistify}{\textsc{DemIstifyCPS}\xspace}

\begin{document}

\title{A Conceptual Framework for Refining Influence Knowledge from Simulation Evidence in Cyber-Physical Systems}

\author{Barbara da Silva Oliveira}
\email{barbara.da-silva-oliveira@univ-cotedazur.fr}
\orcid{0009-0007-8455-9627}
\affiliation{%
  \institution{Université Côte d’Azur,\\ I3S/INRIA Kairos}
  \city{Sophia Antipolis}
  \country{France}
}

\author{Julien Deantoni} \orcid{0000-0001-6962-7846}
\email{julien.deantoni@univ-cotedazur.fr}
\affiliation{%
  \institution{Université Côte d’Azur,\\ I3S/INRIA Kairos}
  \city{Sophia Antipolis}
  \country{France}}

\author{Nicolas Ferry} \orcid{0000-0003-2036-0508}
\email{nicolas.ferry@univ-cotedazur.fr}
\affiliation{%
  \institution{Université Côte d’Azur,\\ I3S/INRIA Kairos}
  \city{Sophia Antipolis}
  \country{France}
}


\begin{abstract}

Cyber-physical systems (CPS) are typically developed by multiple stakeholders who produce artefacts tailored to their specific domains of expertise. The behaviour of these systems emerges from the interaction between those artefacts and their operational environment. Simulation and co-simulation have become essential approaches for analysing CPS behaviour and, through simulation campaigns, developers can explore system responses under changing conditions, including interactions with the environment. However, the lack of details and understanding of some environment-mediated interactions (typically the ones beyond direct sensing and actuation), which remain unmodelled due to their complexity, a lack of time, or a lack of domain experience, hinders the proper comprehension and exploitation of simulation results. To address these limitations, we propose a conceptual framework leveraging the novel concept of \textit{Influences} to support the iterative and incremental refinement of simulation campaigns and deepen the understanding of the system behaviour. We demonstrate the proposed approach through a case study involving a mobile robot implemented using Simulink/Gazebo co-simulation. 

\end{abstract}





\maketitle

\section{Introduction}
\label{introduction}
\input{sections/new_intro}

\section{Background and Related Work}
\label{related}

\input{sections/related}

\section{Motivating Example}
\label{motivating-example}
\input{sections/motivatingExemple}

\input{sections/proposal_JD}

\section{Evaluation}
\label{evaluation}
\input{sections/new_evaluation}
\section{Discussion}
\label{discussion}
\input{sections/discussion}

\section{Conclusion}
\label{conclusion}
\input{sections/conclusion}

\bibliographystyle{ACM-Reference-Format}
\bibliography{acmart}

\end{document}

%% file: sections/new_intro.tex
Cyber-physical systems (CPS) are typically developed by multiple stakeholders who create artefacts according to their domains of expertise. This makes CPS development a socio-technical endeavour, where system integration depends on the coordination of decisions across heterogeneous artefacts \cite{carreira2020multi,combemale2014globalizing}. A significant challenge in CPS development, however, is that their behaviour is not just the sum of isolated components.  Instead, CPS behaviour emerges not only from couplings \textit{within} the System Under Study (SUS) but also from couplings \textit{between} the SUS and the environment in which the CPS operates~\cite{lee2008cyber}. While couplings such as those involving sensing and actuation are generally well-represented, others remain unmodelled due to complexity, a lack of time, or a lack of domain experience. In particular, environment-mediated interactions are often overlooked despite their critical role in shaping system outcomes, thereby significantly limiting the understanding of the system behaviour.

Simulation and co-simulation have become essential approaches for analysing CPS behaviour. Co-simulation enables the joint execution of heterogeneous models developed using different tools, making it possible to study emergent system behaviour across software and physical domains~\cite{gomes2018co}. Through simulation campaigns, developers can explore system responses under varying conditions, including interactions with the environment. Nevertheless, the lack of details and understanding regarding environment-mediated interactions hinders the proper exploitation of simulation results. For instance, requirement violations may appear and stakeholders may not comprehend why they occur and how to coordinate actions to maintain their satisfaction. To improve simulation and make well-justified decisions, stakeholders require a structured way to understand their impact on system behaviour, and to coordinate decisions among themselves. 

To address these limitations, we propose to leverage the concept of \textit{Influences} to guide the simulation of CPS and deepen the understanding of the system behaviour. Influence models and their supporting modelling infrastructure were introduced in prior work \cite{da2026demistifycps}. An \textit{influence} is a modelling concept that captures how design artefacts and environmental factors jointly affect System Response Properties (SRPs), \ie, observable system outcomes such as braking distance or tracking error that ultimately determine requirement satisfaction. Developers may express influences with only a partial understanding of them, for example, only identifying their participants. In this paper we propose a conceptual framework for the iterative and stepwise refinement of simulation campaigns on the basis of influence models with the aim of better formalising and understanding the role of the environment-mediated interactions in the system behaviour. First, Influences are used as a guide by identifying relevant artefacts, environmental factors, and SRPs, thereby structuring and constraining simulation campaigns. Second, they serve as an actionable modelling artefact for knowledge, where evidence extracted from simulation traces is used to refine, extend, and revise influence relationships. In summary, the main contributions of the paper are:
\begin{itemize}
\item \textbf{An influence-guided simulation methodology}, that leverages influence models to design and structure simulation campaigns, extract System Response Properties from execution traces, and align simulation evidence with model elements.
\item \textbf{Simulation-driven refinement of influence models}, to systematically update and extend influence relationships based on simulation data, improving model definitions and deriving lightweight functional abstractions.
\end{itemize}

We demonstrate the proposed approach through a case study involving a mobile robot implemented using Simulink/Gazebo co-simulation. The results show that structuring simulation evidence through influence models improves the understanding of system behaviour and supports more informed design decisions.

The remainder of the paper is organized as follows. Section \ref{related} presents the foundations and related work of this paper. Section \ref{motivating-example} introduces the running example that we use in the paper. Section \ref{sec:approach} presents the methodology for influence-guided simulation. Section \ref{evaluation} presents the evaluation results. Section \ref{discussion} discusses the perspectives and limitations of this work. Section \ref{conclusion} concludes the paper and presents future work.

%% file: sections/related.tex
Model-Based Systems Engineering (MBSE) is a central paradigm for the development of Cyber-Physical Systems (CPS), supporting system specification, verification, and validation through structured and interconnected models~\cite{estefan2007survey,carroll_systematic_2016}. In CPS development, multiple stakeholders contribute models at different levels of abstraction, each reflecting specific domain expertise and concerns. As a result, system-level properties and requirement satisfaction depend not only on individual components, but on the integration and interaction of heterogeneous system artefacts.

Additionally, because CPS operate in a physical environment, system behaviour emerges from interactions both within the System Under Study (SUS) and between the SUS and its surrounding environment~\cite{lee2008cyber,burns2020deriving,ALY2025100044}. Existing modelling approaches capture parts of these interactions. For instance, approaches such as ARCADIA represent external entities as actors that exchange functions and data with the system~\cite{bonnet2017modeling,GARCIAVALLS2018559}. Similarly, context and parametric diagrams allow engineers to represent environmental assumptions and dependencies within system models~\cite{daun20023}. These approaches highlight the importance of explicitly modelling environment-related knowledge through well-defined interfaces, typically associated with sensor inputs and actuator outputs.

However, these modelling techniques primarily focus on direct, interface-based interactions and are less suited to capturing indirect, environment-mediated couplings that influence system behaviour beyond explicit data exchanges. Such interactions arise when environmental factors and system artefacts jointly affect system-level outcomes in ways that are not directly observable through interfaces alone. These interactions can be conceptualized as \textit{influences}, which capture how environmental factors and design artefacts jointly affect system behaviour and, ultimately, requirement satisfaction.

To analyse system behaviour prior to deployment, simulation and co-simulation have become essential in CPS engineering. Co-simulation enables the joint execution of heterogeneous models developed using different tools, making it possible to study emergent system behaviour across software and physical domains~\cite{gomes2018co}. Through simulation campaigns, developers can explore system responses under varying conditions, including interactions with the environment.

Several approaches exploit simulation traces to support the analysis and understanding of system behaviour. Specification-based monitoring techniques verify execution traces against formal properties~\cite{bartocci2018specification}, while other approaches focus on diagnosis, explanation, or root-cause analysis of undesired behaviours~\cite{Jha2023Actionable,maeyens2020process}. In parallel, design space exploration methods use simulation campaigns to explore parameter configurations, often with optimization objectives~\cite{saxena2010mde, beyer2007robust}. Related work on sensitivity analysis and uncertainty quantification, typically implemented through simulation, quantify how variations in input parameters affect system outputs~\cite{Iooss2015}. Additionally, validity frames characterize the conditions under which simulation models provide valid representations of the system, making explicit how environmental conditions bound simulation results~\cite{van2024validity}.

Beyond CPS-specific approaches, several related paradigms address the representation and learning of specific dependencies. Structural Causal Models (SCMs) provide a rigorous mathematical framework for representing cause-effect relationships through directed acyclic graphs (DAGs) and structural equations, enabling interventional analysis and counterfactual reasoning \cite{pearl2009causal}. However, their application in CPS engineering is often challenged by a ``cold start'' problem: constructing the underlying causal structure typically requires either substantial domain expertise or sufficient observational/interventional data for causal discovery, both of which may be limited, especially in a CPS setting. 
In the early design stages of a CPS, the underlying dependency graph is often precisely what is missing or incomplete. Our work differs from SCMs in its engineering-centric focus: rather than performing causal inference on a fixed or latent model, we propose an incremental process where \textit{influence models} serve as a semi-formal scaffold. These models allow engineers to bridge the gap between intuitive domain knowledge and formal dependency structures. In this sense, our approach serves as a domain-specific abstraction that structures the refinement of environment-mediated dependencies; it provides the qualitative and semi-quantitative groundwork necessary to eventually transition toward fully analytical models as the system architecture matures.

Similarly, digital twin and hybrid modelling approaches integrate simulation and data to continuously update system representations, often focusing on runtime monitoring and prediction \cite{edtRoadmap,batty2018digital}. In contrast, our work targets design-time model refinement, with an emphasis on structuring knowledge. 

System identification and data-driven modelling approaches aim to derive system representations from data, typically focusing on estimating parameters or learning functional relationships \cite{ljung1998system}. Our work is complementary, as it operates at the level of structured models and focuses on refining and extending relationships between model elements. While our current implementation relies on lightweight analysis techniques, it can be extended with more advanced learning methods.

Finally, research on model evolution and model repair addresses how models can be incrementally updated to maintain consistency and correctness \cite{mens2006detecting}. These approaches are typically driven by structural constraints and transformations. In contrast, our approach leverages simulation evidence to drive the refinement and completion of behavioural relationships within models.

Despite these advances, existing approaches typically treat simulation outputs as analysis artefacts rather than as inputs for systematically refining system models. In particular, they do not provide mechanisms to structure simulation evidence in a way that supports the incremental enrichment of models.

In contrast, this work leverages simulation campaigns not for fault remediation or optimization, but to generate structured evidence that can be fed back into design-time models, and specifically influence models. The objective is to refine and extend influence models by systematically incorporating knowledge derived from simulation, thereby improving the representation of system behaviour and its underlying dependencies.

%% file: sections/motivatingExemple.tex
In the following, we present a sign-following robot use case, which illustrates current challenges in CPS development and simulation approaches and serves as a running example throughout the paper.

\subsection{Description of Use Case}

We consider the development of an autonomous sign-following robot based on the TurtleBot3 Waffle Pi platform. The robot operates in an indoor space and must navigate a path by detecting and reacting to visual signs indicating left turns, right turns, and a final stop. The system is developed using a combination of domain-specific models and validated through Simulink/Gazebo co-simulation~\cite{MathWorksSignFollowing}.

The system must satisfy the following requirements, expressed in terms of observable system response properties (SRPs):

\begin{itemize}
    \item \textbf{RQ1: Sign Search completion time:} The time spent by the robot to look for a new sign should be below 4 seconds.
    \item \textbf{RQ2: Safe travel speed:} The robot must maintain an average speed of at least 0.05 m/s during forward motion while ensuring that its maximum speed does not exceed 1 m/s.
    \item \textbf{RQ3: Reliable Sign Tracking:} The robot must reliably detect and track signs, maintaining a root-mean-square (RMS) tracking error (compared to centred tracking) below 150 pixels.
\end{itemize}

These requirements are evaluated through SRPs such as average and maximum speed, time to detect a new sign, and tracking error, which characterise some observable behaviours of the system.


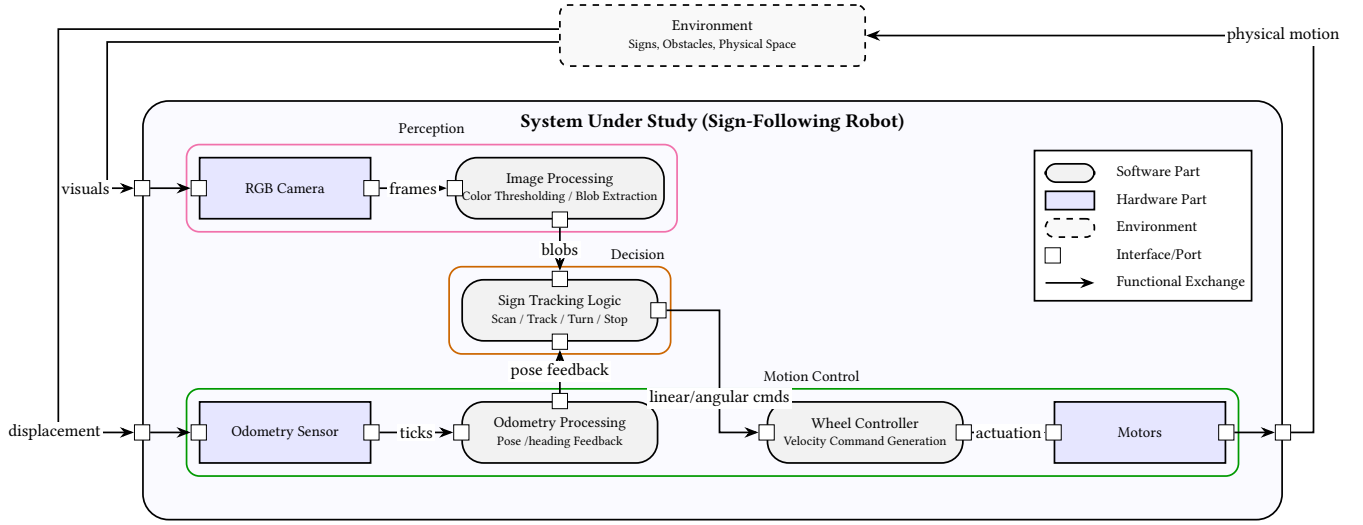
\begin{figure*}[htbp!]
  \centering
  \resizebox{\linewidth}{!}{
    \begin{tikzpicture}[
        >=Stealth,
        node distance=12mm and 10mm,
        every node/.style={font=\footnotesize},
        blk/.style={draw, rounded corners=10pt, thick, fill=gray!10, align=center, minimum width=32mm, minimum height=10mm},
        hw/.style={draw, rounded corners=0pt, thick, fill=blue!10, align=center, minimum width=28mm, minimum height=10mm},
        env/.style={draw, dashed, rounded corners=4pt, thick, fill=gray!05, align=center, minimum width=50mm, minimum height=10mm},
        grp/.style={draw, rounded corners=6pt, thick, inner sep=4mm},
        port/.style={draw, fill=white, minimum size=2.5mm, inner sep=0pt},
        arr/.style={->, thick},
        lab/.style={midway, fill=white, inner sep=1pt, font=\small}
    ]

    \node[env] (world) at (7, 5.5) {Environment\\{\scriptsize Signs, Obstacles, Physical Space}};

    \node[hw] (camera) at (0, 3) {RGB Camera};
    \node[blk] (imgproc) at (4.5, 3) {Image Processing\\{\scriptsize Color Thresholding / Blob Extraction}};

    \node[blk] (decision) at (4.5, 1.0) {Sign Tracking Logic\\{\scriptsize Scan / Track / Turn / Stop}};

    \node[hw] (odom) at (0, -1.0) {Odometry Sensor};
    \node[blk] (odomproc) at (4.5, -1.0) {Odometry Processing\\{\scriptsize Pose /heading Feedback}};
    \node[blk] (wheel) at (9.5, -1.0) {Wheel Controller\\{\scriptsize Velocity Command Generation}};
    \node[hw] (motors) at (14, -1.0) {Motors};

    \node[port] (p_cam_in) at (camera.west) {}; 
    \node[port] (p_odm_in) at (odom.west) {};    
    \node[port] (p_cam_out) at (camera.east) {};
    \node[port] (p_img_in)  at (imgproc.west) {};
    \node[port] (p_img_out) at (imgproc.south) {};
    \node[port] (p_dec_top) at (decision.north) {};
    \node[port] (p_odm_out) at (odom.east) {};
    \node[port] (p_orp_in)  at (odomproc.west) {};
    \node[port] (p_orp_out) at (odomproc.north) {}; 
    \node[port] (p_dec_bot) at (decision.south) {};
    \node[port] (p_dec_out) at (decision.east) {};  
    \node[port] (p_whl_in)  at (wheel.west) {};
    \node[port] (p_whl_out) at (wheel.east) {};
    \node[port] (p_mot_in)  at (motors.west) {};
    \node[port] (p_mot_out) at (motors.east) {};

    \node[grp, draw=magenta!70, fit=(camera)(imgproc), label=above:Perception, inner sep=2mm] (g1) {};
    \node[grp, draw=orange!80!black, fit=(decision), label=above right:Decision, inner sep=2mm] (g2) {};
    \node[grp, draw=green!60!black, fit=(odom)(odomproc)(wheel)(motors), label=above right:Motion Control, inner sep=2mm] (g3) {};

    \begin{scope}[on background layer]
        \node[draw, rounded corners=12pt, thick, fill=blue!02, 
              fit=(g1)(g2)(g3), inner sep=7mm] (sys) {};
        \node[anchor=north, font=\bfseries] at ([yshift=-1mm]sys.north) {System Under Study (Sign-Following Robot)};
    \end{scope}

    \node[port] (b_cam_in) at (p_cam_in -| sys.west) {};
    \node[port] (b_odm_in) at (p_odm_in -| sys.west) {};
    \node[port] (b_mot_out) at (p_mot_out -| sys.east) {};

    
    \draw[arr] ([yshift=-3pt]world.west) -- ++(-7.4,0) |- (b_cam_in) node[lab, pos=0.6, left] {visuals};
    \draw[arr] (b_cam_in) -- (p_cam_in);

    \draw[arr] ([yshift=3pt]world.west) -- ++(-8.2,0) |- (b_odm_in) node[lab, pos=0.8, left] {displacement};
    \draw[arr] (b_odm_in) -- (p_odm_in);

    \draw[arr] (p_cam_out) -- (p_img_in) node[lab] {frames};
    \draw[arr] (p_img_out) -- (p_dec_top) node[lab] {blobs};
    \draw[arr] (p_odm_out) -- (p_orp_in) node[lab] {ticks};
    \draw[arr] (p_orp_out) -- (p_dec_bot) node[lab] {pose feedback};
    \draw[arr] (p_dec_out) -| ++(1, -1.3) |- (p_whl_in) node[lab, pos=0.1] {linear/angular cmds};
    \draw[arr] (p_whl_out) -- (p_mot_in) node[lab] {actuation};

    \draw[arr] (p_mot_out) -- (b_mot_out);
    \draw[arr] (b_mot_out) -| ++(0.5, 6.5) -- (world.east) node[lab, pos=0.2, right] {physical motion};

    \node[draw, thick, fill=white, anchor=south east, xshift=-5mm, yshift=-33mm] (legend) at (sys.north east) {
      \renewcommand{\arraystretch}{1.6}
      \begin{tabular}{@{}ll@{}}
        \tikz[baseline=-0.5ex]{\node[draw, rounded corners=5pt, thick, fill=gray!10, minimum width=8mm, minimum height=3mm, anchor=center] (char) {};} & Software Part \\
        \tikz[baseline=-0.5ex]{\node[hw, minimum width=8mm, minimum height=3mm, anchor=center] (char) {};} & Hardware Part \\
        \tikz[baseline=-0.5ex]{\node[env, minimum width=8mm, minimum height=3mm, anchor=center] (char) {};} & Environment \\
        \tikz[baseline=-0.5ex]{\node[port, anchor=center] (char) {};} & Interface/Port \\
        \tikz[baseline=-0.5ex]{\draw[arr] (0,0) -- (0.8,0);} & Functional Exchange \\
      \end{tabular}
    };

    \end{tikzpicture}
  }
  \caption{High-Level Sign-Following Robot's Architecture.}
  \label{fig:sign_following_high_level_block_diagram}
\end{figure*}


Figure~\ref{fig:sign_following_high_level_block_diagram} depicts the high-level architecture of the system, including software components (\eg, image processing, decision logic, control), hardware elements (\eg, camera, motors, odometry), and their interactions with the environment through sensing and actuation interfaces. At this stage, system models explicitly represent \textit{(i)} functional exchanges between components and \textit{(ii)} sensing and actuation interfaces. 


During development, simulation is used to observe system behaviour under different configurations. The control and decision logic are implemented in Simulink, while Gazebo emulates the robot and its physical environment, with time-synchronized co-simulation enabling integrated execution.

\subsection{Problem Illustration}
\label{motivating-example-problem}

At design time, the system is specified using structured models (\eg, SysML), where interactions between components and with the environment are captured through sensing and actuation interfaces. For example, the perception component processes camera inputs to estimate the position of a visual marker, the decision component determines control actions, and the motion controller executes these commands. Environmental factors such as $AmbientLighting$ and $GroundFriction$ are represented implicitly through input signals or modelling assumptions.

However, a key challenge arises during simulation: the observed system behaviour cannot be fully explained by these interface-level interactions alone. Instead, several \textit{environment-mediated couplings} emerge, where System Response Properties (SRPs) depend on combinations of design artefacts and environmental factors across multiple domains. For instance, the sign detection rate is affected not only by perception parameters such as $BlobSize$, but also by the environmental factor $AmbientLighting$.   
Similarly, the time required to complete the mission depends not only on the nominal velocity of the robot, but also on perception quality and tracking accuracy. These relationships span multiple subsystems and are not explicitly represented in the initial design models. Other examples of such environment-mediated couplings in our motivating example include:
\begin{itemize}
    \item The combined effect of $vNominal$ and $GroundFriction$ on the average speed of the robot.
    \item The influence of $AmbientLighting$, $BlobSize$ and $WallTrans\-parency$ on the sign detection rate.
    \item The interaction between perceived $BlobSize$ and robot dynamics (controller gain) influencing tracking stability.
    \item The joint impact of $vNominal$, $GroundFriction$ and controller gain on mission timing performance.
\end{itemize}
Addressing this challenge requires overcoming two distinct obstacles: ($O_1$) identifying these environment-mediated couplings and their participants; and ($O_2$), defining these couplings to understand their impact on the SRPs and enable downstream design-time analysis.
In current practice, these two aspects are largely handled informally. Simulation campaigns generate large volumes of execution traces, from which engineers must manually infer how specific parameters and environmental conditions affect system behaviour. This knowledge is typically retained as tacit expertise and is rarely formalised, making it difficult to reuse or systematically leverage across analyses. Moreover, this process is time-consuming, error-prone, and lacks systematic support for validating the structure of influences or refining their associated functions as additional knowledge becomes available.

\textbf{This motivates the need for a framework that explicitly links simulation and modelling in a closed loop, incrementally conceptualizing environment-mediated couplings by structuring simulation evidence.}

%% file: sections/proposal_JD.tex
\section{Influence-Guided Simulation and Model Refinement}
\label{sec:approach}

We propose a framework that integrates influence modelling, statistical design of experiments, and sensitivity analysis in a closed-loop process to iteratively refine and complete system knowledge in Cyber-Physical Systems (CPS).
The approach takes as input an initial influence model, typically incomplete and partially specified, and produces a refined model, in which, both the set of relevant participants and their quantitative influence functions are progressively improved using simulation evidence.
%

In the following, we recall the concept of influence modelling, and we detail the overall approach before digging into the key steps it encompasses.

\subsection{Influence Modelling in a nutshell}
\label{influence}
\input{sections/influence}

\subsection{Overall Approach}
\label{overal}

The key idea behind our conceptual framework is to treat the influence model not only as a representation of system knowledge, but as a driver for focused simulation campaigns and as a structured repository for extracted evidence.

The application of the framework requires two inputs: (i) one or more System Response Properties of interest, together with validated procedures to extract these properties from simulation traces; (ii) a bounded participant space defining the simulation exploration domain, including admissible ranges for their variation. Together, these inputs prevent the impractical exploration of all possible design artifacts and environmental factors.

\begin{figure*}
    \centering
    \includegraphics[width=0.93\textwidth]{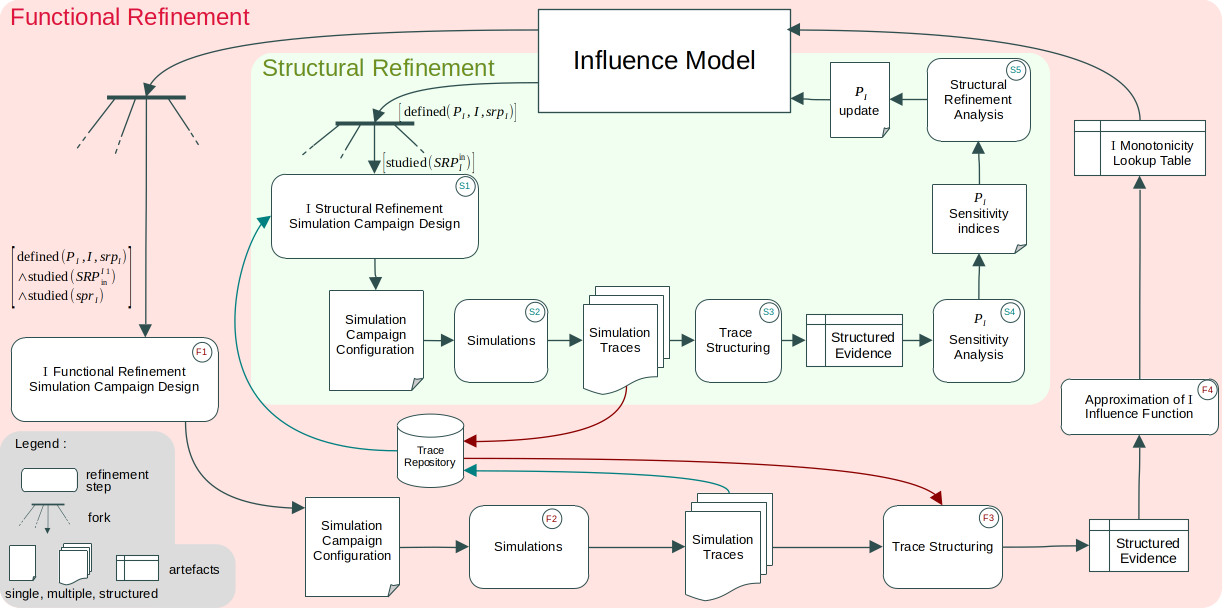}
    \caption{Overview of the approach and its structural and functional refinement loops.}
    \label{fig:overview}
\end{figure*}

As depicted in Figure \ref{fig:overview}, the refinement process comprises two coupled iterative loops: a \textit{structural refinement loop} followed by a \textit{functional refinement loop}. The first loop employs sensitivity analysis to validate and adjust the participants set of each influence. Participants that do not significantly shape the SRP are removed, while additional elements (\ie, Design artefacts $d \in \mathcal{D}$ or environmental factors $\phi \in \Phi$) may be identified and added to the influence participants (thereby addressing obstacle $O_1$). Additionally, the sensitivity indices produced by this loop, which characterise participant importance and their mutual couplings, are also stored in the influence model and used by the second loop to guide the design of simulation campaigns and the refinement of influence function abstractions. The second loop focuses on refining the influence function $f_I$ into a quantitatively analysable artefact, which can be used for downstream analysis and decision-making (thereby addressing $O_2$).

For each influence, separately, the structural refinement loop consists of the following steps:

\begin{enumerate}
    \item[($S_1$)] \textbf{Structural Refinement Simulation Campaign Design:} It results in a set of simulation runs designed to isolate the effect of each participant on the output SRP.
    \item[($S_2$)] \textbf{Simulation Execution and Trace Collection:} The designed campaign is executed and the resulting execution traces are collected.
    \item[($S_3$)] \textbf{Trace Structuring:} the collected traces are processed to construct a structured dataset suitable for analysis.
    \item[($S_4$)] \textbf{Sensitivity Analysis:} The collected data is analysed to compute sensitivity indices that quantify the influence of each participant on the output SRP, as well as the presence of interactions between participants.
    \item[($S_5$)] \textbf{Structural Refinement Analysis:} The sensitivity indices are used to validate (or invalidate) the relevance of each participant, to identify potential missing participants. It produces updated participant sets and sensitivity indices.
\end{enumerate}

It is worth noting that influences can have SRPs as participants. In such case, because the observed range of the output SRP is used to define the input space for the next influence, the structural refinement of upstream influences needs to be performed first.

Once the structure of the influence is revised, the functional refinement loop is triggered to refine the influence function $f_I$ that describes how the participants co-affect the SRP. This loop consists of the following steps:

\begin{enumerate}
    \item[($F_1$)] \textbf{Functional Refinement Simulation Campaign Design:} The design of the simulation campaign is informed by the sensitivity indices obtained during structural refinement, which guide the selection of parameter combinations to explore for capturing interactions and non-linear effects.
    \item[($F_2$)] \textbf{Simulation Execution and Trace Collection:} The designed campaign is executed and the resulting execution traces are collected.
    \item[($F_3$)] \textbf{Trace Structuring:} The collected traces are processed to construct a structured dataset suitable for analysis.
    \item[($F_4$)] \textbf{Abstraction of Influence Functions:} The dataset is used to abstract the influence function $f_I$, resulting in a quantitatively analysable artefact.
\end{enumerate}

Note that all simulation traces are stored in a repository where traces can be retrieved as needed to avoid redundant simulation efforts.

Once the influence function is refined, the resulting model is proposed to designers (\eg, via version-controlled pull requests), enabling the incremental update of system knowledge based on empirical evidence; enabling downstream design-time analysis to guide decision-making and socio-technical considerations. In this process, simulation is not just a validation tool but a systematic means to construct, validate, and refine system knowledge.

Each step of each of these loops is detailed in the following subsections.

\subsection{The Structural Refinement Loop}
\label{sec:structural-ref}

Before any simulation campaign can be launched, the system must be instrumented to capture the relevant System Response Properties (SRPs) from simulation logs.
Two complementary strategies are employed. The first involves configuring the simulation infrastructure to log SRP values directly during execution. This requires prior identification of the relevant log streams, system variables, or output files. The second, when SRPs are not directly available, requires  explicit extraction procedures to derive them from raw simulation traces. This involves parsing logs, selecting relevant events, and applying aggregation or filtering functions. Examples of extraction functions include filtering events by type or timestamp range, aggregating metrics (\eg, maximum latency, mean throughput) or computing derived quantities (\eg, cumulative resource usage).

This extraction layer must be validated against domain knowledge to ensure that computed SRPs faithfully represent the intended system behaviour. Once extraction procedures are established and validated, they are applied uniformly across all simulation runs in the campaign.

\subsubsection{Design of Simulation Campaign for Structural Refinement}
\label{sec:exploratory_screening}

Simulation campaigns are derived from the currently established influence model. The structure of the influence is used to decompose the global simulation problem into smaller, influence-specific sub-campaigns, \ie, one campaign per influence. For a given influence $I$, the union of participants $P_I = D_I \cup \Phi_I \cup \mathcal{SRP}^{\mathrm{in}}_I$ defines the input space for the campaign, while $\mathrm{srp}_I$ defines the output to be observed. This enables an exploration focused on specific relations and to design experiments that are targeted at validating and refining the influence model. Note that because the participants and SRP of an influence are involved, this structure should be defined prior to analysis.

The goal of each campaign is to assess, rank and refine the participants of a specific influence. There exist many methods related to sensitivity analysis that can be used to select the actual configuration. They range from the simplest such as OFAT (One-Factor-At-a-Time) to more advanced methods such as Sobol's indices~\cite{saltelli2010variance}. All of them are compatible with our approach, but they require different amounts of experiments. In this paper, the Morris method~\cite{morris1991factorial} is selected due to its efficiency in the number of required simulations. This method constructs a set of trajectories within the input space. In each trajectory, one participant is varied at a time by a predefined step, while the others participants are fixed. In this way, it isolates the effect of a variation in one participant under a particular configuration of the remaining participants. The output of this process is a configuration file that defines the parameters to be varied, their range of variation, and the sampling strategy. This configuration is then used to execute the simulation campaign.

To identify whether additional participants should be included into an influence, engineers define a bounded candidate set $C_I$ containing design artefacts and environmental factors which are not currently included in $P_I$. At selected points in the current exploration domain, we systematically vary those candidates. This ``extended'' Morris screening detects whether variations in these external factors produce significant changes in the output SRP. A candidate is then proposed for addition in the influence definition when its estimated effect exceeds the uncertainty threshold introduced in Section \ref{subsec:structural-analysis}. This procedure can identify omitted participants in the bounded set $C_I$, it does not search the unrestricted CPS design and environment space.


\subsubsection{Simulation Execution and Trace Collection}
Once the campaigns are specified for each influence, the simulation executes each run, \ie, execution of one simulation for each configuration variant as defined in the previous stage. Each run produces a trace that includes the configuration variants and participant settings. The trace is assumed to be a chronologically ordered sequence of events, each carrying a timestamp and an optional value. In co-simulation settings,  an orchestration mechanism must ensure synchronization between simulators, covering time-step coordination and data exchange policies. The trace can then be post-processed to extract and attach relevant SRP values according to the previously established procedures. To account for stochasticity, repeated simulations are performed at selected design points to estimate aleatory uncertainty. This aleatory uncertainty ($\sigma_{nat}$) serves as a baseline for subsequent analysis.

\subsubsection{Trace Structuring and Dataset Construction}

The simulation campaigns produce execution traces that are typically unstructured and high-dimensional. To extract meaningful insights, we apply a systematic process to structure these traces into datasets that align with the influence model. This structured representation, which is common to both loops, enables further analysis.

Simulation outputs are transformed into structured datasets where each record corresponds to a simulation run and includes:
\begin{itemize}
    \item the set of participant configurations (\ie, values of $D_I$, $\Phi_I$, and $\mathcal{SRP}^{\mathrm{in}}_I$)
    \item the value of the output SRPs (and specifically the value $\mathrm{srp}_I$) and their uncertainty estimates (\eg, mean and variance from repeated runs)
    \item metadata linking the record to the specific influence, model version, a campaign identifier and simulator settings, and a random seed (including randomized run ordering).
\end{itemize}

To ensure traceability and avoid redundant computations, all records are additionally added to a central repository. This repository can be queried to retrieve specific influence runs that could be reused, reducing the computational load of the campaign induced by simulation. For instance, runs used during the structural refinement can also be exploited during the functional refinement.

\subsubsection{Sensitivity Analysis for Structural Refinement}

This step aims at analysing the structured dataset to provide markers that identify which participants have a significant influence on the output SRP, and which can be considered negligible. To achieve this, we apply the Morris elementary effects method~\cite{morris1991factorial}. This global sensitivity analysis technique offers a way to screen for influential factors in high-dimensional models and is particularly suitable for our context due to its ability to provide both qualitative and semi-quantitative insights at a relatively low computational cost.
For each participant, the simulation outputs are used to compute elementary effects, which measure how the output SRP changes according to a predefined variation of that participant. This method provides two primary indices: the mean absolute elementary effect ($\mu^*$), which represents the overall importance of each participant on $srp_I$; and the standard deviation ($\sigma$), which captures the variability of its effects across the explored domain. A high $\sigma$ indicates the presence of non-linear effects, interactions between participants, or both.

As an output of this step, we rank participants by their significance and characterise the interactions between them for the specific influence under study. These insights inform the subsequent structural refinement, which is detailed in the next section.

\subsubsection{Structural Refinement Analysis}
\label{subsec:structural-analysis}
The objective of the structural refinement analysis is to exploit the sensitivity indices to validate and update the set of participants $P_I$ associated with an influence. This step operationalizes the transition from simulation evidence to model structure. Based on the sensitivity analysis, the decision to prune or not negligible participants is taken as follows.

%
%
Participants with high $\mu^*$ are considered influential, as they induce significant variations in the output SRP and therefore constitute the core of the influence structure. Conversely, participants with low $\mu^*$ have a limited impact on $srp_I$ over the explored domain and can be considered for removal. This distinction enables a principled simplification of the influence by focusing only on the most impactful contributors.
To ensure robustness, this pruning decision is not based solely on relative rankings but is grounded in the aleatory uncertainty estimated during the simulation campaign. Specifically, currently, we consider a participant as negligible if its estimated contribution to output variation, approximated by $\mu^*$ scaled by the Morris step size, remains below the baseline variability $\sigma_{nat}$, meaning that its influence cannot be distinguished from stochastic noise. Different strategies may be considered.
Beyond pruning, the sensitivity analysis also supports the identification of missing participants. During the exploratory screening phase, factors not initially included in $P_I$ are perturbed (see Section~\ref{sec:exploratory_screening}). If such a factor exhibits a significant $\mu^*$ (\ie, above the uncertainty-informed threshold $\sigma_{nat}$), it is flagged as a relevant contributor and incorporated into the influence model. This mechanism enables identifying omitted candidates within the explored space and mitigates the risk of incomplete influence specifications.

The result of this step is thus a refined set of participants annotated with their relative importance (thereby addressing $O_1$). The estimated $\mu^*$ values can be normalised and directly stored as weights in the influence model, providing a first-order quantitative characterisation of participant importance, as leveraged in the design-time analysis described in~\cite{da2026demistifycps}. Additional information regarding interaction effects is preserved for the design of the functional refinement simulation campaign described in the next section.

\subsection{The Functional Refinement Loop}

This loop focuses on refining the influence function $f_I$ that describes how the participants co-affect the output SRP. The goal is to derive a quantitatively analysable artefact that captures how variations in the participants affect the system outcome, which can be used for downstream analysis and decision-making. This process is guided by the sensitivity indices obtained during structural refinement, which inform the design of targeted simulation campaigns to capture non-linearities and interactions, as well as the selection of appropriate function approximation techniques. It assumes a sufficiently refined structure from the previous loop, as the presence of irrelevant participants can obscure the functional relationships and lead to misinterpretation of the influence function.

The subsequent sections detail the design of the simulation campaign for functional refinement and the abstraction of influence functions into a quantitatively analysable form. Simulation and trace structuring are common to both loops and are described in the previous sections.

\subsubsection{Simulation Campaign Design for Functional Refinement}

The goal of this stage is to devise further experiments with the objective to collect sufficient data for characterising the influence function. The campaign design is guided by the results of the sensitivity analysis. In particular, two indicators are used. The mean absolute elementary effect, \(\mu^*\), estimates the overall importance of a participant, it is used to inform the structure of the experimental design. The standard deviation of its elementary effects, \(\sigma\), indicates how much its effect varies across the explored input space, and it is used to govern the exploration resolution.



High $\sigma$ values for a subset of variables suggest the presence of significant interactions or non-linear behaviours. In such cases, the simulation campaign must jointly vary these participants to capture their coupled effects. This typically requires factorial or space-filling designs over the corresponding subspace. 
A factorial design systematically evaluates combinations of specified participant levels, but its cost increases very rapidly depending on the number of participants and levels. Space-filling designs are a more scalable option, distributing samples in the input space, which helps reducing the cost of experiments. In this context, we employ a Latin Hypercube Sampling (LHS) strategy \cite{McKay1979}, which produces combinations of the input dimension that cover the exploration domain without the need to calculate every configuration.

Low $\sigma$ values suggest that the model response can be approximated as predominantly additive. In such cases, simpler and more cost-effective experimental designs, such as One-Factor-At-a-Time (OFAT) strategies, which independently estimate the contribution of each variable while reducing computational cost, can be adopted.

Once the structure of the design is defined, the relative importance values are leveraged to control the resolution of the exploration. Participants with a strong effect on the output SRP are assigned finer discretizations (\ie, a higher number of levels). Less influential participants are assigned coarser discretizations or are fixed to nominal values when their contribution is negligible. 

The $\mu^*$ must be normalised so that they can be compared among participants. For each participant $i$, let $\mu_i^*$ denote the Morris importance obtained in the corresponding upstream structural campaign. Two normalised forms are computed:
\begin{equation}
w_i^{\max} = \frac{\mu_i^*}{\max_j \mu_j^*},
\qquad
w_i^{\sum} = \frac{\mu_i^*}{\sum_j \mu_j^*}.
\end{equation}
The first normalisation is used to parameterize the Weighted Latin Hypercube Sampling (WLHS), while the second is kept as a normalised importance score for reporting. 

When the distribution of the importance weights $\mu^*$ is heterogeneous and $\sigma$ is high, a Weighted Latin Hypercube Sampling (WLHS) strategy is adopted to efficiently explore the input space. While standard LHS enforces a uniform distribution across all dimensions~\cite{McKay1979}, WLHS adapts the sampling density according to the relative importance of each variable~\cite{WLHS}. This enables a more refined exploration of influential variables and their interactions by allocating simulation effort according to their weights, thereby improving the efficiency of the campaign while maintaining adequate coverage of the input space.

Overall, the structural refinement directly informs both the organization and the resolution of the experimental design, ensuring that the simulation campaign focuses on the most critical regions of the input space.

\subsubsection{Functional Approximation of Influence Functions}

Once the simulation campaign is completed and the raw traces are structured, the resulting datasets are used to approximate the influence function $f_I$. The objective is to derive a compact representation of how the participants in $D_I$, $\Phi_I$, and $\mathcal{SRP}^{\mathrm{in}}_I$ affect the output $\mathrm{srp}_I$. Instead of fitting a full regression or surrogate model, we construct an intermediate abstraction that is lightweight and directly suited to design-time storage and reasoning.

In this work, this abstraction takes the form of a so-called \emph{Monotonicity Lookup Table}. For each participant, the table captures its piecewise monotonic behaviour over input value ranges. This is achieved by recording, from input value ranges automatically extracted from the simulation traces, the local slope values of the variation of the output SRP with respect to variations in the participant. In other words, it indicates how much increasing the participant within a given value range results into an increasing or decreasing trend in the output SRP. Because this relationship may not be globally monotonic, the table encodes these slope patterns separately across relevant input value intervals automatically identified from the simulation data.
This range-based representation provides a semi-quantitative characterization of how each participant influences the system output, while avoiding the data requirements and computational complexity of full regression-based approaches.
Note that when the interaction indicator $\sigma$ is significant for a subset of participants, this suggests the presence of strong interactions or non-linear behaviours. In such cases, a purely univariate representation may be insufficient to fully capture the system dynamics. Conceptually, the proposed framework extends to a conditional monotonicity representation, in which the monotonicity of a given participant is evaluated with respect to its value range \emph{and} conditioned on the ranges of other participants. This extension enables the capture of joint effects between inputs by explicitly accounting for how the influence of one participant depends on the state of others.

%% file: sections/influence.tex
The concept of Influence was introduced to capture how design artefacts and environmental factors jointly shape system behaviour, which cannot be defined by standard interface contracts or dependencies. An influence, as represented in Figure \ref{fig:influenceDef}, explicitly connects participants (design artefacts and/or environmental factors) to an output System Response Property (SRP), and requirements can be expressed as constraints or margins over SRPs.

\begin{figure}[htbp!]
\centering
\usetikzlibrary{shapes.geometric, positioning, arrows.meta, patterns, calc} 
\resizebox{\linewidth}{!}{
    \begin{tikzpicture}[
    font=\sffamily,
    node distance=1.5cm,
    environment_sphere/.style={
        circle,
        minimum size=1.5cm,
        draw=blue!80!black,
        very thick,
        pattern=horizontal lines,
        pattern color=blue!60!cyan,
        fill=blue!5!white,
        inner sep=0pt
    },
    design_sphere/.style={
        circle,
        minimum size=1.5cm,
        draw=green!60!black,
        very thick,
        pattern=grid,
        pattern color=green!40!lime,
        fill=green!5!white,
        inner sep=0pt
    },
    SR_sphere/.style={
        circle,
        minimum size=1.5cm,
        draw=red!60!black,
        very thick,
        pattern=grid,
        pattern color=red!40!lime,
        fill=red!5!white,
        inner sep=0pt
    },
    influence/.style={
        rectangle,
        draw=black,
        thick,
        fill=white,
        minimum width=2.5cm,
        minimum height=1cm,
        align=center,
        font=\small\sffamily
    },
    system_response/.style={
        ellipse,
        draw=purple!80!black,
        thick,
        fill=purple!20!white,
        minimum width=2.5cm,
        minimum height=1cm,
        align=center,
        font=\small\sffamily
    },
    requirement/.style={
        rectangle,
        draw=black,
        thick,
        dashed,
        fill=white,
        minimum width=2.5cm,
        minimum height=1cm,
        text width=2.3cm,
        align=center,
        font=\small\sffamily
    },
    arrow/.style={
        -Stealth,
        thick,
        draw=black
    },
    dashed_arrow/.style={
        -Stealth,
        thick,
        draw=black,
        dashed
    },
    compensating_arrow/.style={
        -Stealth,
        thick,
        draw=cyan!70!blue,
        dashed,
        bend left
    }
]

\coordinate (envcoord) at (0,0) ;
\node[environment_sphere, pattern color=blue!70!cyan, fill=blue!10] at ($(envcoord.center) + (-0.2, 0.2)$) {};
\node[environment_sphere, pattern color=blue!50!cyan, fill=blue!10!white] at ($(envcoord.center) + (-0.1, 0.1)$) {};
\node[environment_sphere] (envmain) at (envcoord) {};
\node[left=0.25cm of envmain.west, font=\sffamily\small\bfseries, align=center] (envlabel) {Environmental \\ Factors ($\Phi_{I}$)};

\coordinate[design_sphere] (descoord) at (0,-4.2); 
\node[design_sphere, pattern color=green!30!lime, fill=green!10!white] at ($(descoord.center) + (-0.2, 0.2)$) {};
\node[design_sphere, pattern color=green!30!lime, fill=green!10!white] at ($(descoord.center) + (-0.1, 0.1)$) {};
\node[design_sphere] (desmain) at (descoord) {};
\node[left=0.25cm of desmain.west, font=\sffamily\small\bfseries, align=center] (deslabel) {Design Artefacts \\ ($\mathcal{D}_{I}$)};

\coordinate[SR_sphere] (SRcoord) at (0,-2.1); 
\node[SR_sphere, pattern color=red!30!lime, fill=red!10!white] at ($(SRcoord.center) + (-0.2, 0.2)$) {};
\node[SR_sphere, pattern color=red!30!lime, fill=red!10!white] at ($(SRcoord.center) + (-0.1, 0.1)$) {};
\node[SR_sphere] (SRmain) at (SRcoord) {};
\node[left=0.25cm of SRmain.west, font=\sffamily\small\bfseries, align=center] (SRlabel) {System Response \\ Properties ($\mathcal{SRP}^{in}_{I}$)};
\node[influence, right=of $(envmain.east)!0.5!(desmain.east)$] (inf) {Influence ($I$) \\ $\mathrm{srp}_I$ = f($\Phi_{I}$,$\mathcal{SRP}^{in}_{I}$, $\mathcal{D}_{I}$)};
\node[system_response, right=of inf] (sysresp) {System \\ Response\\ Property ($\mathcal{\mathrm{srp}_I}$)};

\node[requirement, above right=0.76cm and -0.24cm of sysresp] (reqbehind1) {};
\node[requirement, above right=0.68cm and -0.32cm of sysresp] (reqbehind2) {};
\node[requirement, above right=0.6cm and -0.4cm of sysresp] (req) {};
\node[above=0.1cm of req.north, font=\sffamily\small\bfseries] (reqlabel) {Requirements ($\mathcal{R}$)};

\path[arrow] (envmain.east) -- ($(inf.west)+ (0,0.25)$);
\path[arrow] (SRmain.east) -- ($(inf.west)- (0,0)$);
\path[arrow] (desmain.east) -- ($(inf.west)- (0,0.25)$);
\path[arrow] (inf) -- (sysresp);
\path[dashed_arrow] (sysresp) -- (req);

\path[compensating_arrow, bend left=-35] ($(sysresp.north)+(-0.6,-0.1)$) to ($(envmain.east)+(-0.1,0.20)$);
\path[compensating_arrow, bend right=-35] ($(sysresp.south)+(-0.6,0.1)$) to ($(SRmain.east)+(-0.1,-0.20)$);
\path[compensating_arrow, bend right=-35] ($(sysresp.south)+(-0.3,0.06)$) to ($(desmain.east)+(-0.1,-0.20)$);

\node[draw, align=left, font=\small, anchor=south east, fill=white] at ($(sysresp.south east)+(2.5,-2)$) (legend) {
    \begin{tikzpicture}[baseline={(0,0)}]
        \path[compensating_arrow] (0,0) -- (0.8,0);
        \node[right=2pt] at (0.8,0) {Affected by};
        \path[dashed_arrow] (0,-0.5) -- (0.8,-0.5);
        \node[right=2pt] at (0.8,-0.5) {Constrained by};
    \end{tikzpicture}
};

\end{tikzpicture}
}

\caption{Illustration of the concept of Influence \cite{da2026demistifycps}.}
\label{fig:influenceDef}
\end{figure}

The work \cite{da2026demistifycps} proposes a modelling framework consisting of a DSL integrated within the Eclipse Modeling Framework (EMF~\cite{EMF}), and therefore compatible with languages such as SysML, together with an analysis engine that produces design-time feedback in the form of model annotations. In this framework, an \textbf{Influence} is defined as: ``An \emph{Influence} is a relation that explicitly connects multiple participants, such as design artefacts or environmental factors, to a system response property. By definition, each influence includes at least one environmental factor or system response property among its participants. Variations in the system response property may directly or indirectly affect the satisfaction of requirements. Influences can be specified at different levels of abstraction, and may be composed to capture complex interactions. Each influence provides: \textit{(i)} explicit identification of all participants; \textit{(ii)} a system response property; \textit{(iii)} a qualitative or quantitative description of how participant changes affect the system response property; and \textit{(iv)} traceability, either directly or through chains of influences, to the related requirements, supporting analysis, decision-making, and cross-domain coordination''. Formally, an influence instance $I$ is defined as the tuple: 
\begin{equation}
I \;=\; \bigl(D_I,\; \Phi_I,\; \mathcal{SRP}^{\mathrm{in}}_I,\; \mathrm{srp}_I,\; f_I\bigr),
\end{equation}

where:
\begin{itemize} 
  \item $D_I \subseteq \mathcal{D}$ is the (nonempty) set of design artefacts participating in $I$. Thus $D_I \neq \varnothing$.
  \item $\Phi_I \subseteq \Phi$ is the set of environmental factors participating in $I$. $\Phi_I$ may be empty only if $\mathcal{SRP}^{\mathrm{in}}_I \neq \varnothing$. 
  \item $\mathcal{SRP}^{\mathrm{in}}_I \subseteq \mathcal{SRP}$ is the (possibly empty) set of SRPs that act as inputs to $I$. $\mathcal{SRP}^{\mathrm{in}}_I$ may be empty only if $\Phi_I \neq \varnothing$.
  \item $\mathrm{srp}_I \in \mathcal{SRP}$ is the SRP that is affected (the output) of $I$,
  \item $f_I$ is the influence function describing how the participants co-affect the value of $\mathrm{srp}_I$.
\end{itemize}

\begin{figure*}[h!]
\centering
\includegraphics[width=0.8\linewidth]{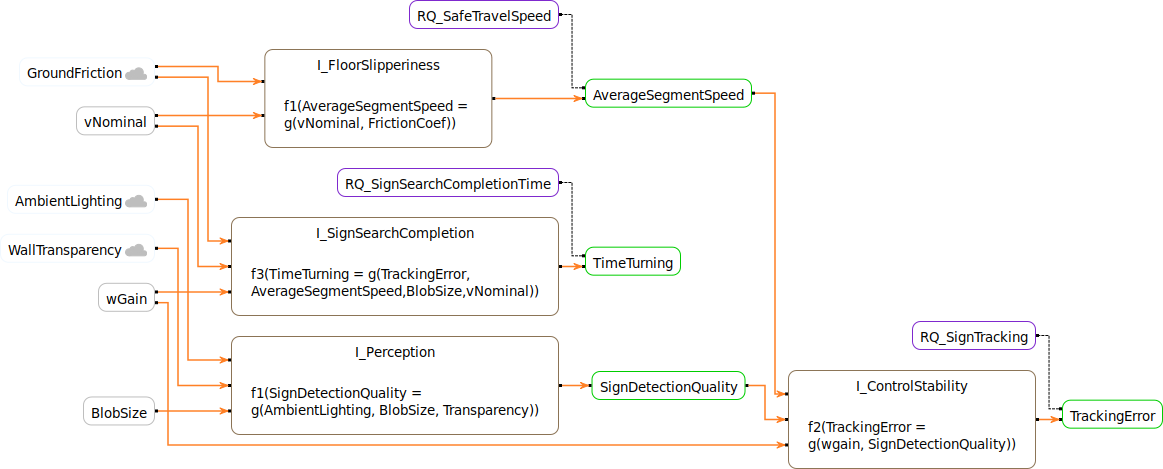}
\caption{Diagram of the initial influences modelled with \demistify DSL using KIELER Lightweight Diagrams.}
\label{fig:influencesDiagram}
\end{figure*}

Figure \ref{fig:influencesDiagram} shows the initial Influence Model
of the running example introduced in Section \ref{motivating-example}. The model is specified using the \demistify DSL and visualised using KIELER Lightweight Diagrams \footnote{\url{https://github.com/kieler/KLighD}}. It contains four influences. The Perception influence relates $D_I=\{BlobSize\}$ and $\Phi_I=\{AmbientLighting,WallTransparency\}$ to the output SRP $SignDetectionQuality$. The Floor Slipperiness influence relates $D_I=\{vNominal\}$ and $\Phi_I=\{GroundFriction\}$ to $AverageSegmentSpeed$. The Control Stability influence relates $D_I=\{wGain\}$ and $\mathcal{SRP}^{\mathrm{in}}_I= \{SignDetectionQuality,AverageSegment\-Speed\}$ to $TrackingError$. Finally, the Sign Search Completion influence relates $D_I=\{wGain,vNominal\}$ and $\Phi_I=\{GroundFriction\}$ to $TimeTurning$.

The figure also illustrates influence composition:
$SignDetection\-Quality$ and $AverageSegmentSpeed$ are both outputs of
upstream influences and inputs of the Control Stability influence.

This structure makes explicit which elements participate in each influence. However, structural information does not explain how changes in these participants affect the output SRP. This knowledge can be represented using the so called influence function $f_I$. 

The influence function should not be interpreted as a single fixed mathematical object, but rather as a concept that can be instantiated at different levels of abstraction. At a coarse level, the function may be informally described (\eg, in natural language), capturing qualitative relationships between participants and the output SRP. At more refined levels, it may be specified as a semi-formal or fully formal mathematical function, enabling explicit evaluation of how variations in participants affect the SRP.

This multi-level nature implies that different levels of analysis and reasoning can be performed depending on the available information about the influence function. For instance, when only structural information is available (\ie, the sets of participants), the framework defined in~\cite{da2026demistifycps} can analyse the potential impact of changes to design artefacts or environmental factors by examining their connectivity within the influence model. This supports, among other analysis, the identification of stakeholders and artefacts potentially affected by a given change.

As the influence function becomes more precisely specified, additional reasoning becomes possible. With a more detailed function definition, the framework can assess how changes in a participant affect the output SRP, including whether such changes are beneficial or prejudicial with respect to requirement satisfaction. This enables the identification of trade-offs between stakeholders and supports informed decision-making and socio-technical coordination. It is important to note there is a \emph{minimal level of precision} at which an influence enables the full set of analysis methods defined in~\cite{da2026demistifycps}. At this level, the influence function provides sufficient structure to characterize the relative importance (or weight) of participants within the influence; and determine how variations in each participant propagate to the output SRP.
Below this minimal level, an influence remains useful for traceability and structural reasoning, but does not support quantitative or impact-based analysis.

%% file: sections/new_evaluation.tex
The co-simulation campaign was executed on an HP EliteBook 840 G10 workstation equipped with a 13th Gen Intel Core i7-1370P processor and 64 GB of RAM, running Ubuntu 22.04.5 LTS. The models and control logic were executed using MATLAB/Simulink (2024a~\cite{MATLAB:2024}), interfaced via ROS2~\cite{macenski2022ros2} with Gazebo~\cite{koenig2004gazebo}. The Gazebo simulation and TurtleBot3 physical kinematics are hosted within an Ubuntu Focal 20.04 Virtual Machine. The experiments and related code are available in the repository: \url{https://github.com/barbara-da-silva-oliveira/simulation-driven-influence-refinement}.

The evaluation is structured around the two obstacles introduced in the paper in Section \ref{motivating-example-problem}: ($O_1$) assessing whether the framework supports the definition and refinement of influences and their participants, and ($O_2$) assessing whether the framework enables the derivation and refinement of functional abstractions suitable for downstream design-time reasoning.

\subsection{Structural Refinement Results}

This section presents the results of the structural refinement loop (Section~\ref{sec:structural-ref}) applied to the initial influence model of our running example shown in Figure~\ref{fig:influencesDiagram}. The goal is to validate and refine the participants of each influence by removing non-relevant elements and characterizing their relative importance and interactions. Each influence is analysed independently following the defined process.



\begin{table}[h!]
\caption{Structural refinement results.\\ {\tiny(C.: Campaign; $\boldsymbol{\mu^*}$: mean absolute elementary effect; $\boldsymbol{\sigma}$: standard deviation; $\boldsymbol{\sigma_{nat}}$:  aleatory uncertainty; Dec.: Decision on participant)}.}
\label{tab:structural-results-summary}
\centering
\small
\begin{tabular}{llcccl}
\toprule
\textbf{C.} & \textbf{Participant} & $\boldsymbol{\mu^*}$ & $\boldsymbol{\sigma}$ & $\boldsymbol{\sigma_{nat}}$ & \textbf{Dec.} \\
\midrule
P & AmbientLighting    & 0.0120 & 0.0230 & 0.0048 & Retain \\
P & BlobSize      & 0.2270 & 0.0055 & 0.0048 & Retain \\
P & WallTransparency  & 0.0071 & 0.0120 & 0.0048 & \textbf{Prune} \\
\midrule
S & $v_{\mathrm{nominal}}$ & 0.1522
 & 0.02461 &  0.0042& Retain \\
S & GroundFriction           & 0.03102
 & 0.05659 & 0.0042 & Retain \\
\midrule
C & wGain $w_{\mathrm{gain}}$         & 25.2755 & 35.9776

&  3.9485 &  Retain\\
C & SignDetectionQuality &
81.03846 & 59.5928
& 3.9485 & Retain\\
C & AverageSegmentSpeed &  22.1525& 35.2820
& 3.9485&  Retain\\
\midrule
M & $v_{\mathrm{nominal}}$    & 1.8
 & 1.6120
 & 0.6789 & Retain \\
M & GroundFriction      & 2.4375
 & 3.8433
 & 0.6789 & Retain \\
M & wGain      & 2.3063
 & 3.5377
 & 0.6789 & Retain \\

\bottomrule
\end{tabular}       
\end{table}

\subsubsection{Campaign $P$: Perception Influence}

Campaign $P$ targets the Perception influence, relating \textit{AmbientLighting}, \textit{BlobSize}, and \textit{WallTransparency} to the output SRP \textit{SignDetectionQuality}, which is the proportion of frames where a sign is detected. The simulation campaign comprised 16 design points with two replicates each (\ie, 32 runs). 
As reported in Table~\ref{tab:structural-results-summary}, the Morris analysis identifies \texttt{BlobSize} as the dominant participant (highest $\mu^*$), followed by \textit{AmbientLighting}, while \textit{WallTransparency} has a lower influence. The $\sigma$ values indicate non-additive effects: \textit{AmbientLighting} and \textit{WallTransparency} exhibit noticeable dispersion, whereas \textit{BlobSize} combines high importance with lower interaction effects. 
%
From a structural perspective, the outcome is not only a ranking of participants but also a decision on their inclusion in the influence. \textit{WallTransparency} is the weakest participant, and its estimated effect magnitude ($\mu^* . \Delta x = 0.00071$) is much lower than the aleatory baseline ($\sigma_{nat}=0.0048$). As its influence on the SRP cannot be clearly distinguished from noise, it is considered negligible and pruned from the influence model. 

\subsubsection{Campaign $S$: Floor Slipperiness Influence}

Campaign $S$ targets the Floor slipperiness influence, relating \textit{vNominal} and \textit{GroundFriction} to the output SRP \textit{AverageSegmentSpeed} in m/s. The simulation campaign comprised 12 design points with two replicates each (\ie, 24 runs). The resulting structured data shows an \textit{AverageSegmentSpeed} ranging from 0.1866 to 0.3401 over the explored domain.
As reported in Table~\ref{tab:structural-results-summary}, the Morris analysis identifies \textit{vNominal} as the dominant contributor ($\mu^*=0.1522$), while \textit{GroundFriction} has a secondary effect ($\mu^*=0.0310$). For \textit{GroundFriction}, the dispersion is larger than its mean effect ($\sigma=0.0566 > \mu^*$), indicating significant non-linearities and/or interaction effects, and suggesting this influence cannot be adequately captured by a purely additive model. From the structural point of view, both participants must be retained.

\subsubsection{Campaign $C$: Control-Stability Influence}

Campaign $C$ analyses the \textit{Control-Stability} influence by studying how \textit{TrackingError} varies with the controller gain \textit{wGain}, under the $\textit{I\_Perception}$ and $\textit{I\_Slippery}$ upstream SRPs. The \textit{Sign\-De\-tec\-tion\-Quality} and \textit{AverageSegmentSpeed} define the conditioning context.
Based on upstream results, their values were partitioned into `Low', `Medium', and `High' ranges. Co-occurring values were grouped into representative joint regimes (\eg, `High–Low', `Low–High', `Low–Medium', `Medium–Low', `Medium–Medium'), capturing empirically observed system states.
For each regime, \textit{wGain} was varied to analyse its effect on \textit{TrackingError}, ensuring exploration grounded in observed upstream behaviour rather than artificial combinations. Two representative scenarios were retained per non-empty regime, except for `Medium–Medium' (only one instance available), resulting in 27 design points (\ie, 54 runs).
As reported in Table~\ref{tab:structural-results-summary}, the Morris analysis identifies \textit{SignDetectionQuality} as the dominant contributor ($\mu^*=81.0385$), while \textit{AverageSegmentSpeed} has the least effect ($\mu^*=22.1525$). From the structural point of view, all participants must be retained.

\subsubsection{Campaign $M$: Sign Search Completion Influence}

Campaign $M$ targets the \textit{I\_Sign-Search-Completion} influence, relating \textit{wGain}, \textit{vNominal}, and \textit{GroundFriction} to the output SRP \textit{TimeTurning} in seconds, which is the time the robot spends on turning to find a new sign. The simulation campaign comprised 16 design points with two replicates each (\ie, 32 runs).
As reported in Table~\ref{tab:structural-results-summary}, the analysis identifies \textit{GroundFriction} as the dominant participant (highest $\mu^*$), followed by \textit{wGain}, while \textit{vNominal} has a lower influence. The $\sigma$ values suggest non-linearities and interaction effects, indicating the presence of nonlinear behaviour and/or interactions over the explored domain. All participants exhibit effects above the aleatory uncertainty threshold and are therefore retained in the influence model.




\subsection{Functional Refinement}
\label{sec:FuncRef}

In this section we overview the two main steps of the functional refinement, \ie the simulation campaign design and the approximation of the influence function.

Whilst the functional refinement was applied to all the influences, for the sake of space, we only report here the results for the \textit{Floor Slipperiness} influence. The structural refinement indicated that the corresponding influence function cannot be adequately captured by a purely additive model. Consequently, a Weighted Latin Hypercube Sampling (WLHS) strategy was used to design the simulation campaign, resulting in 16 design points (\ie, 32 runs).


Once the simulations were executed (step F2 in Figure~\ref{fig:overview}) and the traces structured (step F3 in Figure~\ref{fig:overview}), the influence function approximation is computed, resulting in the monotonicity lookup tables shown in Table~\ref{S_FUNC}.

\begin{table}[h]
\centering
\begin{tabular}{cccc}
\hline
\textit{GroundFriction} & \textit{vNominal} & \textbf{SRP local slope} \\ \hline
$[0.25, 0.34]$ & $[0.2707, 0.3312]$ & 0.7803 \\ \hline
$[0.38, 0.50]$ & $[0.2529, 0.3516]$ & 0.8062 \\ \hline
$[0.52, 0.57]$ & $[0.2804, 0.3254]$ & 0.8902 \\ \hline
\end{tabular}
\caption{\textit{AverageSegmentSpeed} monotonicity lookup table.}
\label{S_FUNC}
\end{table}

The results indicate that the influence function is globally monotonic: \textit{AverageSegmentSpeed} increases with the most influential participant, \textit{vNominal}, over the explored domain (i.e., the estimated SRP local slope is consistently positive). 

Furthermore, the local slope of the SRP with respect to \textit{vNominal} depends on the friction coefficient. To capture this dependency, the \textit{GroundFriction} domain is partitioned into intervals with comparable sample sizes, and the SRP local slope is estimated within each interval. The results show that the influence of \textit{vNominal} on the SRP increases with the friction coefficient over the explored range.

This observation is consistent with the sensitivity analysis, where \textit{GroundFriction} exhibited a dispersion larger than its mean effect ($\sigma = 0.0566 > \mu^*$), indicating non-linearities and interaction effects. It also aligns with the physical intuition that higher wheel–ground friction enables a more effective transfer of commanded speed to the robot’s actual motion.

The resulting lookup table is stored in the influence model as a representation of the influence function, enabling downstream design-time analysis.

%% file: sections/discussion.tex
The evaluation shows that the proposed framework can turn initially partial influence knowledge into a more explicit and operational representation of CPS behaviour. The key contribution of this work is not only the use of simulation to reason about logs, but also the structuring and exploitation of simulation evidence to better understand system behaviour and refine the definition of influences. The influence model is used as both a guide for specialized campaigns and a repository for the knowledge extracted from them.

With respect to obstacle $O_1$, the results of the structural refinement provide an empirical understanding of the relevance of a participant is in shaping a System Response Property. This brings two main benefits. First, it improves the qualitative understanding of the robot's behaviour by explaining which factors matter for each SRP and how these couplings propagate across influences. Second, it provides the structural basis for the functional refinement loop, where the retained participants and the observed interaction patterns are used to design richer campaigns. The present evaluation demonstrates pruning and ranking, however, it does not empirically demonstrate the ability to add candidates from a bounded set.

With respect to obstacle $O_2$, the functional refinement results show that the influence function can be abstracted into a lightweight design-time artefact. The framework uses information obtained from the sensitivity analysis of the structural refinement phase to drive Weighted Latin Hypercube Sampling, concentrating exploration effort on the most influential regions of the input space. The resulting Monotonicity Lookup Table captures how the system response properties vary according to different configurations of participants. This is meaningful in the present case study, where for campaign $S$, we were able to extract both the expected global monotonicity of the influence function and the non-linearity suggested during the sensitivity analysis. These results support the claim that the framework refines not only influence structure, but also the semantic content of influence functions for later design-time enrichment and reasoning.


A threat to validity lies in the dependence of the approach on the correctness of the extraction of SRPs from simulation logs. A mismatch between the logged variables and the intended behavioural properties may affect the interpretation of influences. In practice, this extraction process should be validated by domain experts to ensure that the computed SRPs faithfully represent the targeted system behaviours.
The results also depend on the selected exploration ranges, the Morris design, and the limited number of replicates used to estimate aleatory uncertainty. These choices may influence the observed sensitivity patterns and limit the precision and generalisability of the analysis. The approach has also applicability boundaries. It can only screen omitted participants when they are included in a bounded candidate set and can be varied in simulation.

Moreover, the current evaluation relies on a single mobile-robot case study. While this use case is appropriate to illustrate how environment-mediated couplings span perception, locomotion, and control behaviours, further validation across diverse CPS domains and simulation toolchains is needed to assess the general applicability of the approach. More evaluations should investigate systems with larger participant spaces, more expensive simulations, and stronger stochastic behaviour. Methods such as regression, surrogate-modelling, or other sensitivity-analysis would also help evaluate the accuracy and cost-effectiveness of the proposed abstractions.



%% file: sections/conclusion.tex
This paper presented a conceptual framework that leverages the concept of \textit{Influences} to (i) drive and refine CPS simulation in the presence of environment-mediated couplings, and (ii) improve the understanding of system behaviour under varying design choices and environmental conditions. This is achieved by integrating influence modelling, simulation campaigns, trace structuring, and sensitivity analysis into a closed-loop process, where simulation results are not only used for validation but also as evidence to progressively refine influence knowledge.

The sign-following robot case study illustrated how the approach supports (i) the identification of environment-mediated couplings and their participants (addressing challenge $O_1$), and (ii) the derivation of lightweight functional abstractions that characterize these couplings and their impact on SRPs, enabling downstream design-time analysis (addressing challenge $O_2$). Overall, the framework enables a transition from incomplete and informal knowledge of environment-mediated couplings toward structured, evidence-based, and analysable representations of system behaviour.

This work also opens several perspectives. In particular, integrating the proposed framework within a Digital Twin setting would enable the continuous incorporation of operational data, allowing influence models to evolve beyond design-time artifacts and reflect changes in environmental conditions over time.